\documentclass[letterpaper, 10 pt, conference]{ieeeconf}  

\IEEEoverridecommandlockouts                              

\title{\LARGE \bf
MFCalib: Single-shot and Automatic Extrinsic Calibration for LiDAR and Camera in Targetless Environments Based on Multi-Feature Edge}

\author{Tianyong Ye$^{1}$, Wei Xu$^{2}$, Chunran Zheng$^{3}$ and Yukang Cui$^{1}$
\thanks{*This work was partially supported by National Key R\&D Program of China (2023YFE0126800), National Natural Science Foundation of China (61903258), Guangdong Basic and Applied Basic Research Foundation (2024A1515030153) and the Project of Department of Education of Guangdong Province (2022KTSCX105, 2023ZDZX4046), Shenzhen-Hong Kong-Macau Technology Research Programme (SGDX20230821091559019), and Shenzhen Natural Science Fund (Stable Support Plan Program 20231122121608001). (\emph{Corresponding author: Yukang Cui.})}
\thanks{$^{1}$T. Ye and Y. Cui are with the College of Mechatronics and Control Engineering, Shenzhen University, Shenzhen, 518060, China, and also with the Peng Cheng Laboratory, Shenzhen 518000, China. (eserda0114@gmail.com, cuiyukang@gmail.com)}%
\thanks{$^{2}$W. Xu is with the Manifold Tech Limited, Hong Kong, China. (buaaxw@gmail.com)}%
\thanks{
$^{3}$C. Zheng is with the department of Mechanical Engineering, The University of Hong Kong, Hong Kong SAR, China.
(zhengcr@connect.hku.hk)}%
}

\usepackage{bm}
\usepackage[utf8]{inputenc}
\usepackage[noend]{algpseudocode}
\usepackage{amsmath}
\usepackage{amssymb}    
\usepackage[utf8]{inputenc}
\usepackage{xcolor}
\usepackage{graphicx}
\usepackage{hyperref}
\usepackage{booktabs}
\usepackage{color}
\usepackage{array}
\usepackage{booktabs}
\usepackage{multirow}
\usepackage{hyperref}
\usepackage[space]{cite}
\usepackage[linesnumbered,boxed,ruled,commentsnumbered]{algorithm2e}
\graphicspath{{figures/}}

\begin{document}

\maketitle
\begin{abstract}

This paper presents MFCalib, an innovative extrinsic calibration technique for LiDAR and RGB camera that operates automatically in targetless environments with a single data capture. At the heart of this method is using a rich set of edge information, significantly enhancing calibration accuracy and robustness. Specifically, we extract both depth-continuous and depth-discontinuous edges, along with intensity-discontinuous edges on planes. This comprehensive edge extraction strategy ensures our ability to achieve accurate calibration with just one round of data collection, even in complex and varied settings. Addressing the uncertainty of depth-discontinuous edges, we delve into the physical measurement principles of LiDAR and develop a beam model, effectively mitigating the issue of edge inflation caused by the LiDAR beam. Extensive experiment results demonstrate that MFCalib outperforms the state-of-the-art targetless calibration methods across various scenes, achieving and often surpassing the precision of multi-scene calibrations in a single-shot collection. To support community development, we make our code available open-source on GitHub.

\end{abstract}

\section{INTRODUCTION}

In recent years, LiDAR and camera sensors have become essential for perception tasks in robotic systems, including Simultaneous Localization and Mapping (SLAM) \cite{ORB-SLAM2,xu2021fast,point-lio,voxel-map} and semantic segmentation \cite{peng2023bevsegformer,ding2023lenet}. By providing rich texture information and three-dimensional measurements of the environment, LiDAR and camera complement each other, making LiDAR-camera fusion a promising approach to enhance the performance of perception systems. Consequently, the development of robust and accurate LiDAR-camera fusion methods 
\cite{zhu2021camvox,lin2021r,zheng2022fast,zheng2024fast,lin2022r} has become increasingly important.

\begin{figure}[t] 
\centering 
\includegraphics[width=1\columnwidth]{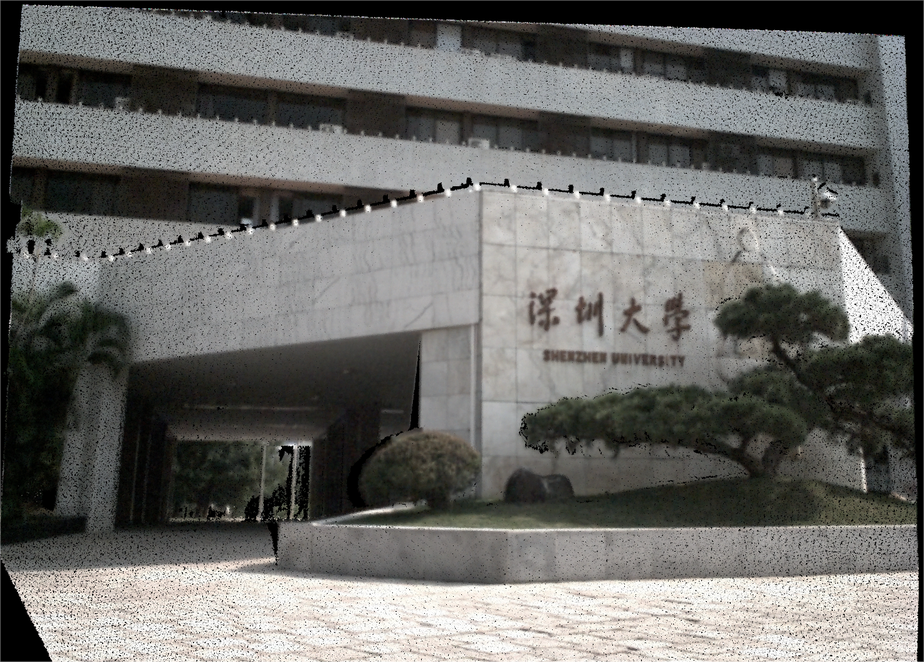} 
\caption{The RGB-colored point cloud using the proposed method.} 
\label{fig:1} 
\vspace{-0.5cm}
\end{figure}

Extrinsic calibration, the estimation of the six degrees-of-freedom (DoF) transformation (rotation and translation) between the LiDAR and camera coordinate systems, is a critical step in sensor fusion. Fig. \ref{fig:1} illustrates an RGB-colored point cloud generated using our proposed method, showcasing the high accuracy of the extrinsic parameters obtained. Among the state-of-the-art extrinsic calibration methods, target-based methods, such as those using checkerboards \cite{beltran2022automatic,cui2020acsc,Zhou_Li_Kaess_2018} or specific image pattern \cite{Chen_Liu_Liang_Zhang_Hyyppa_Chen_2020}, have shown effectiveness and robustness. However, they typically rely on specific calibration targets, limiting their practicality in real-world applications. Targetless calibration methods, which eliminate the need for external calibration targets, often have specific scene requirements \cite{pixel-level,Chen_Li_Zhang_Wu_Wang_2023} and tend to utilize only a single feature from the LiDAR, restricting their robustness in diverse environments.

To mitigate these obstacles, we present MFCalib—a novel method for automatic extrinsic calibration in targetless environments. Our approach combines multi-feature edges extracted from LiDAR point clouds with natural edge features extracted from images through a single data collection, minimizing reprojection errors. Furthermore, we tackle the issue of depth-discontinuous edge inflation in LiDAR by modeling the LiDAR beam. Our method is adaptable to various indoor and outdoor scenes, quickly adjusting extrinsic parameters with just one data collection, overcoming the limitations of target-based approaches.
Our contributions are delineated as follows:

\begin{figure*}[t]
    \begin{center}
        {\includegraphics[width=2.05\columnwidth]{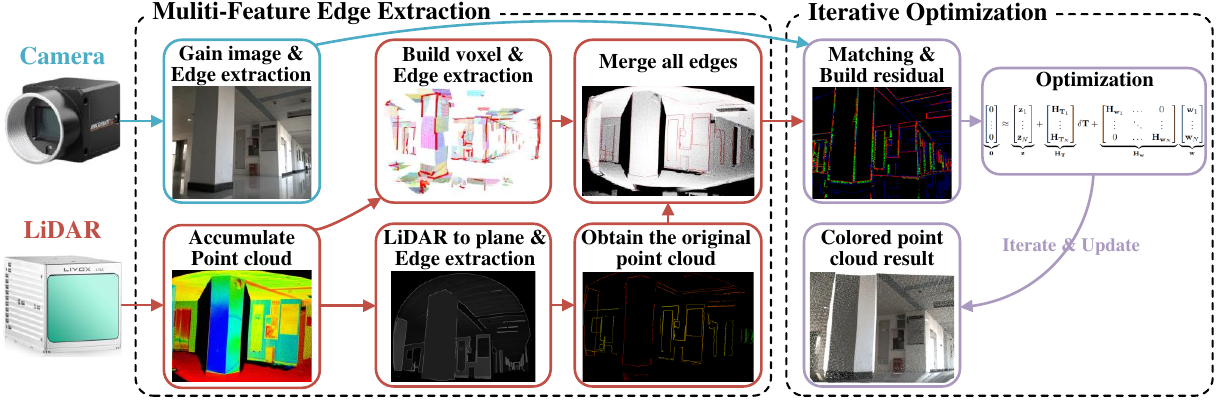}}
    \end{center}
    \vspace{-0.2cm}
    \caption{\label{fig:2}System overview of MFCailb.  }
    \vspace{-0.3cm}
\end{figure*}
\begin{itemize}

\item We introduce an extrinsic calibration strategy that leverages multi-feature: depth-continuous edges, depth-discontinuous edges, and intensity-discontinuous edges within a unified data collection framework. This method enhances robustness by utilizing various types of natural environment edges.

\item In our analysis, we concentrate on the beam divergence angle inherent to LiDAR technology. We model the LiDAR beam and introduce measurement uncertainty for depth-discontinuous edges, effectively addressing the challenge of inaccurately extracting depth-discontinuous edges from the point cloud.

\item Our algorithm, rigorously tested in diverse indoor and outdoor environments, proves to outperform other the state-of-the-art targetless-based methods. It achieves calibration accuracy within a single scene comparable to the cumulative performance of other methods across multiple scenes. To support community development, we make our code available open-source on GitHub\footnote[4]{\url{https://github.com/Es1erda/MFCalib}}.

\end{itemize}

\section{RELATED WORKS}

LiDAR-camera extrinsic calibration methods are primarily categorized into two approaches: 1) target-based and 2) targetless-based. The main distinction between them lies in how they define and extract features from the two sensors.

\subsection{Target-based Calibration}
Target-based methods utilize well-defined objects like checkerboards \cite{beltran2022automatic,cui2020acsc,Zhou_Li_Kaess_2018}, calibration room \cite{Xie_Shao_Guli_Li_Wang_2018} and geometric solids \cite{Kummerle_Kuhner_2020,Park_Yun_Won_Cho_Um_Sim_2014} for accurate transformation estimation. Specifically, Beltrán \textit{\textit{et al.}}\cite{beltran2022automatic} described a process that began with extracting reference points from a custom calibration target based on sensor data, followed by determining the optimal rigid transformation through registration of point sets from both sensors. Xie \textit{et al.} \cite{Xie_Shao_Guli_Li_Wang_2018} developed a calibration space adorned with Apriltags, allowing the calibration of multiple cameras and LiDARs even without a shared field of view (FoV). The positions of cameras and LiDARs were precisely determined using the tags and the room's geometry. Despite their precision, these methods encounter challenges in target design and practical application within dynamic environments.

\subsection{Targetless-based Calibration}

Targetless-based methods leverage natural features from the environment such as planes\cite{chen2022pbacalib,li2023joint}, edges\cite{zhu2021camvox,pixel-level,Liu_Yuan_Zhang_2022,miao2023coarse}, and intensity\cite{koide2023general} variations for extrinsic calibration. Chen \textit{et al.} \cite{chen2022pbacalib} introduced a method based on Plane-Constrained Bundle Adjustment, optimizing extrinsic parameters by iteratively minimizing reprojection errors using feature points obtained from prominent planes in the scene. Introduced in \cite{zhu2021camvox}, LiDAR points are projected onto the image plane, coloring them based on depth and intensity before extracting 2D edges for matching with image edges.. In \cite{Pandey_McBride_Savarese_Eustice_2022}, the authors optimize extrinsic parameters by maximizing the mutual information (MI) between color maps and images. In \cite{pixel-level,Liu_Yuan_Zhang_2022}, natural 3D and 2D edges are extracted from LiDAR point cloud and camera images, respectively; these 3D edges are then projected onto the image plane for optimization. In \cite{koide2023general}, the authors utilize intensity features from the natural environment, employing Superglue\cite{Sarlin_DeTone_Malisiewicz_Rabinovich_2020} for image matching to find 2D-3D relationships between LiDAR and camera and optimizing extrinsic parameters through direct LiDAR-camera registration based on normalized information distance and cross-modal distance metrics using MI. Motion-based methods introduced in \cite{Taylor_Nieto_2016,Liu_Zhang_2021} recover external parameters through sensor self-motion, further refined with appearance information. However, these approaches typically require significant sensor movement to activate calibration fully.

Our method can be considered an enhancement of \cite{pixel-level}, representing a targetless-based methodology. Compared to \cite{pixel-level}, we have several key enhancements:
\begin{itemize}
    \item We employ a multi-feature paradigm that incorporates depth-continuous, depth-discontinuous, and intensity-discontinuous edges, enhancing calibration robustness and accuracy across varied environments.
    \item We tackle the unreliability of depth-discontinuous edge by focusing on LiDAR's beam divergence angle, modeling the beam, and introducing measurement uncertainty, which improves depth-discontinuous edge extraction and calibration reliability.
\end{itemize}

Unlike \cite{zhu2021camvox,Sarlin_DeTone_Malisiewicz_Rabinovich_2020}, we utilize 3D edge features from the point cloud, avoiding the problems associated with multi-valued and zero-valued mapping, thus ensuring our method's adaptability and reliability across various calibration scenarios.

\section{METHODOLOGY}

\subsection{Overview}

Fig. \ref{fig:2} illustrates the comprehensive procedure of the proposed calibration algorithm. Initially, LiDAR is utilized to acquire a dense point cloud, alongside image capture by the camera. The Canny algorithm \cite{canny1986computational} is then employed to extract edges from the image. The extraction of LiDAR edges is divided into two parallel paths: on the one hand, depth-continuous edges are extracted based on building voxel; on the other hand, both depth-discontinuous and intensity-discontinuous edges are extracted using the intensity image from the LiDAR. These extracted edges are then mapped back to the point cloud and merged with all edges. Ultimately, 3D LiDAR edges are projected onto the image plane, constructing residuals for iterative optimization to obtain the result.

\subsection{Initialization}
Calibration starts with the initialization of extrinsic parameters derived from CAD models and the pre-calibration of camera intrinsics. Data collection for LiDAR varies by type: for non-repetitive scanners like Livox Avia\footnote[5]{\url{https://www.livoxtech.com/avia/}}, a static accumulation is used to gather dense cloud. For rotating scanners like Ouster\footnote[6]{\url{https://ouster.com/products/scanning-lidar/os1-sensor/}}, a LiDAR-inertial odometry\cite{fast-lio2} system is applied to accumulate point cloud.

\subsection{Muliti-Feature Edge Extraction}

In the calibration of multi-sensor systems, especially for high-resolution LiDAR, the zero-valued and multi-valued mapping problem\cite{yuan2021arxiv} poses a significant challenge. To circumvent this problem caused by occlusions, our approach involves using the 3D edges extracted from the point cloud in the LiDAR coordinate system and projecting them onto the image in the camera coordinate system to avoid this issue.

In the extrinsic calibration of LiDAR and camera, edge extraction and registration play pivotal roles. Existing methods\cite{pixel-level,Liu_Yuan_Zhang_2022,miao2023coarse} typically rely on a singular type of feature for calibration, which can limit their applicability in complex environments due to the stringent requirements of the environment. To extract as many edge features as possible across a single environment and thus enhance the robustness of the algorithm, we employ the following three distinct types of edge features for registration:

\begin{figure}[htbp] 
\centering 
\includegraphics[width=1.0\columnwidth]{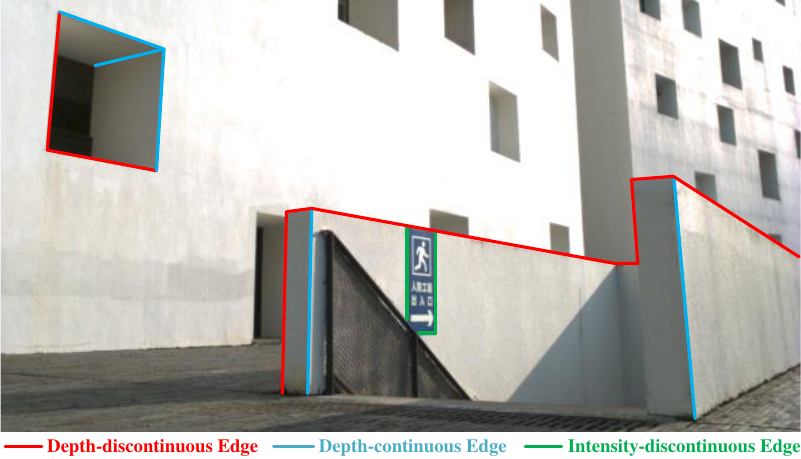} 
\caption{Different kinds of edges in a real-world scene: Red lines mark depth-discontinuous edges, blue lines mark depth-continuous edges, and green lines mark intensity-discontinuous edges on plane.} 
\label{3} 
\vspace{-0.3cm}
\end{figure}

\begin{itemize}
\item Depth-continuous edge: These edges represent the smooth transitions and connections on the surface of objects, but the pixel gradient changes in the image are usually not pronounced.

\item Depth-discontinuous edge: These edges demarcate distinct foreground-background transitions, prominently discernible in imagery. Their alignment enhances depth perception, thus fortifying calibration integrity.

\item Intensity-discontinuous edge: These edges, highlighted by changes in intensity due to material or color variations, unveil the intrinsic properties of the object's surface. Analyzing intensity variations on a uniform plane allows for the effective extraction of these edges.
\end{itemize}

 For the extraction of depth-continuous edges, we adopt the method proposed by Yuan\cite{pixel-level}, which is based on point cloud voxel cutting and plane fitting to extract depth-continuous edges efficiently and accurately.

During the process of extracting both depth-discontinuous and intensity-discontinuous edges, we initially map the data onto a spherical coordinate system to better adapt to the inherent measurement geometry of LiDAR. Subsequently, this spherical representation is converted into a Cartesian coordinate system to produce a 2D intensity image. In this image, each pixel represents a cluster of points in 3D space, with the average intensity value of these points used to color each pixel, while also preserving their depth information. Thereafter, we apply the Canny edge detection algorithm to the intensity map to extract all edges. A filtering mechanism based on KD-Tree is then utilized to differentiate between depth-discontinuous edges and intensity-discontinuous edges on the plane, based on locality and depth variation. The depth-discontinuous and intensity-discontinuous edges extracted in this manner are subsequently re-mapped onto the original LiDAR point cloud.

Within the LiDAR point cloud, we merge all the extracted edges to obtain multi-feature edges. This approach fully leverages the various edge features within the calibration scene and is conducted entirely within the LiDAR coordinate system, avoiding the zero-valued and multi-valued mapping problem. This significantly enhances the robustness and accuracy of the calibration algorithm, thereby making it suitable for a broader range of application scenarios.

\subsection{LIDAR-Camera Registration}
\subsubsection{Edge Matching}\label{edge matching}  
Edge matching and optimization processes are conducted within the image space of the camera coordinate system. Edge points from the LiDAR coordinate system are projected onto the image plane of the camera coordinate system using the formula
\begin{align}
\mathit{{}^C\!}\mathbf{P}_i &= \mathit{{}^C_L\!}\mathbf{T} (\mathit{{}^L\!}\mathbf{P}_i) = \mathit{{}^C_L\!}\mathbf{R}\cdot\mathit{{}^L\!}\mathbf{P}_i + \mathit{{}^C_L\!}\mathbf{t} \in \mathbb{R}^3,
\label{eq:projection}\\
\mathbf{p}_i &= \bm{\pi}({}^C\mathbf{P}_i).
\label{eq:2}
\end{align}

In the above equations, \( {}^C\mathbf{P}_i \) and \( {}^L\mathbf{P}_i \) represent the 3D points in the camera and LiDAR coordinate systems, respectively. Eq. (\ref{eq:projection}) denotes the application of the rigid transformation \( {}^C_L\mathbf{T} \) to the point \( {}^L\mathbf{P}_i \), where \( \mathbf{R}\) is the rotation matrix and \( \mathbf{t}\) is the translation vector. The symbol \( \mathbf{p}_i \in \mathbb{R}^2 \) indicates the location of the point in the camera image space. The function \( \bm{\pi}({}^C\mathbf{P}_i) \) is the camera projection model, which embody the intrinsic transformation from \( {}^C\mathbf{P}_i = [X_i, Y_i, Z_i]^T \) to \( \mathbf{p}_i = [x_i, y_i]^T \).

We initiate a query for the \(\kappa\) nearest neighbors of \(\mathbf{p}_i\) in the KD-Tree generated from the edge pixels of the image. We denote the \(\kappa\) nearest neighbors as \(\mathbf{N}_i = \left\{ \mathbf{q}_i^m; m = 1, \ldots, \kappa \right\}\). The line characterized by \(\mathbf{N}_i\) is determined by the point \(\mathbf{q}_i\) situated on the line itself, accompanied by the normal vector \(\mathbf{n}_i\).


\begin{figure}[t] 
\centering 
\includegraphics[width=1\columnwidth]{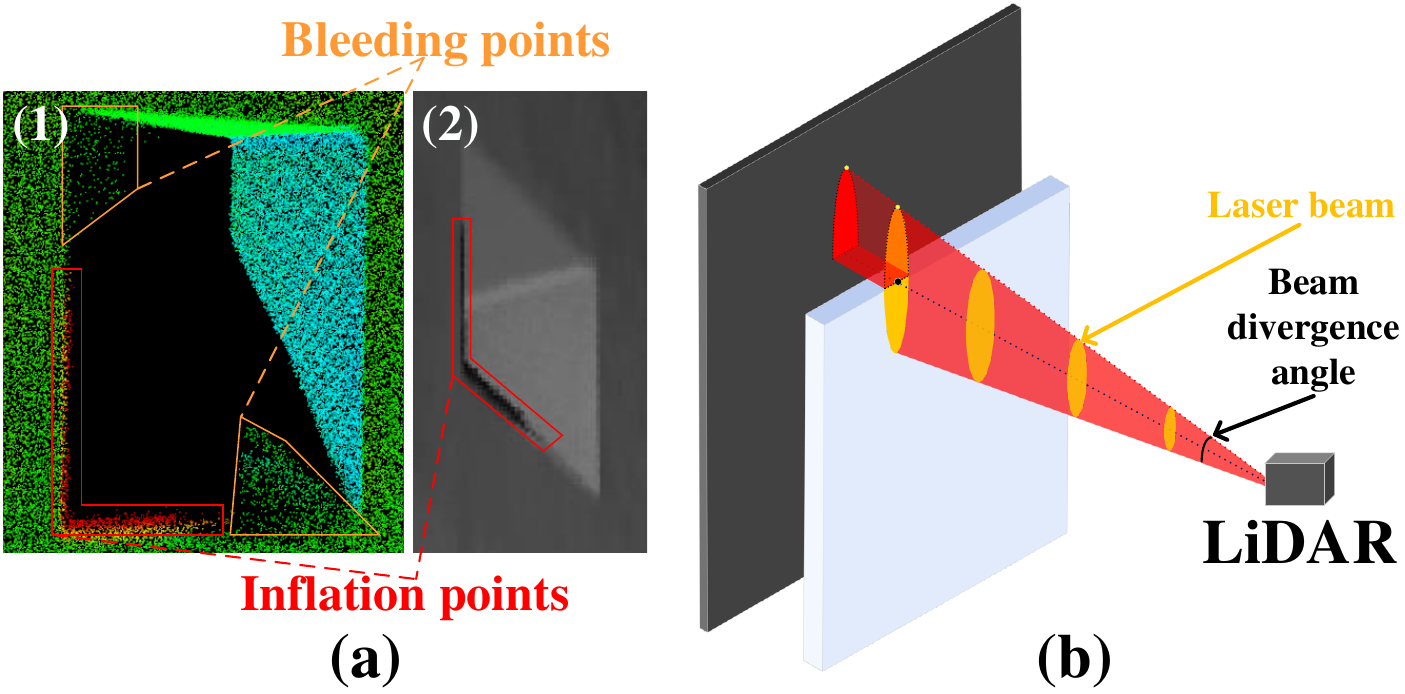} 
\caption{Inflation points and bleeding points caused by the divergence angle of LiDAR laser beam:(a) the actual inflation and bleeding points, where (1) is the point cloud colored by intensity values, and (2) is the intensity image; (b) the schematic of the LiDAR beam. } 
\label{4} 
\vspace{-0.5cm}
\end{figure}

\subsubsection{LiDAR Measurement Noises and Bias}\label{measurement noises}
The LiDAR edge points \(^L\mathbf{P}_i \), that are extracted, along with the corresponding edge features \((
\mathbf{n}_i,\mathbf{q}_i)\) in the image, are subject to the effects of measurement noises. For the inherent measurement noises of LiDAR and camera, we adapt the noise model used in \cite{yuan2021arxiv}. For the camera's measurement noise, the noise \( ^I\mathbf{w}_i \) is modeled as a normally distributed variable \( \mathcal{N}(0, {}^I\mathbf{\Sigma}_{i}) \), which perturbs the edge feature \( \mathbf{q}_i \) during the process of image edge extraction. For points extracted by the LiDAR, the noise associated with these points is denoted by \( ^L\mathbf{w}_i \). The corresponding mathematical expression is given by
\begin{align}
    ^L{\mathbf{P}_{i}^{gt}} \approx  {}^L{\mathbf{P}}_i - {}^L{\mathbf{w}}_i,
\end{align}
where
\begin{align*}
^L{\mathbf{w}}_i &= \mathbf{A}_i \begin{bmatrix}
\delta{d_i} \\
\delta{\bm \omega_i}
\end{bmatrix}
\sim \mathcal{N}(0, {}^L\mathbf{\Sigma}_{i}), \\
^L\mathbf{\Sigma}_{i} &= \mathbf{A}_i 
\begin{bmatrix}
\Sigma_{d_i} & \mathbf{0}_{1 \times 2} \\
\mathbf{0}_{2 \times 1} & \Sigma_{\omega_i}
\end{bmatrix}
\mathbf{A}_i^T,
\end{align*}
in the context of these equations, we derive the relationship between the true position of the point \(^L{\mathbf{P}_{i}^{gt}} \) and its observed counterpart \({}^L{\mathbf{P}}_i\). Here, \( \omega_i\in \mathbb{S}^2 \) denotes the direction of the measured bearing, while \( \delta\omega_i \sim \mathcal{N}(\mathbf{0}_{2 \times 1}, \boldsymbol{\Sigma}_{\omega_i}) \) represents the noise inherent in the measurement within the tangent plane of \( \mathbf{w}_i \), which is assumed to follow a normal distribution. The variable \( d_i \) corresponds to the depth measurement, whereas \( \delta d_i \sim \mathcal{N}(0, \boldsymbol{\Sigma}_{d_i}) \) indicates the corresponding error in ranging, also assumed to be normally distributed. The matrix \( \mathbf{A}_i \) is defined in \cite{yuan2021arxiv}.

Based on this, we further introduce a measurement error caused by the divergence angle of the LiDAR beam. In our simplified model for subsequent optimization, we approximate the shape of the LiDAR beam as a circular form to facilitate easier computation. 
In subsequent optimizations, we consider the measurement errors caused by the divergence angle of the LiDAR beam. Fig. \ref{4}(b) illustrates that laser pulses are not perfect points but rather beams with a divergence angle. Scanning from foreground to background objects results in part of the laser pulse being reflected by the foreground, with the remainder by the background, creating two reflective pulses reaching the LiDAR receiver. When the foreground object is highly reflective, the signal from the first pulse dominates, even when the beam centerline has moved beyond the foreground object. This leads to false points beyond the actual edge of the foreground object. The closer proximity of the foreground to the background merges the signals from both pulses, producing an integrated signal that generates a set of points bridging foreground and background, known as bleeding points, known as bleeding points, shown in Fig. \ref{4}(a). Inflation points result in anomalous intensity values on depth-discontinuous edges, affecting their precision and causing the edges to appear inflated, as observable in Fig. \ref{4}(a). While bleeding points are also anomalies, they have almost no impact on the intensity image of the LiDAR.

The uncertainty of depth-discontinuous edges caused by inflation points due to the LiDAR beam adheres to a biased Gaussian distribution \((\mu, ^L\mathbf{\Sigma}_{i}) \). When the LiDAR beam's centerline is precisely on the depth-discontinuous edge, the measured value is the depth of the foreground object, which is the critical position for phantom point generation. As the centerline of the beam moves up, if the foreground object remains within the range of the beam, ideally, the measurement should return the value of the background object reached by the centerline, but due to the measuring principle of LiDAR, the foreground object's value is still returned. When the elliptical beam is exactly tangent to the depth-discontinuous edge (see Fig. \ref{4}), the LiDAR still measures the value of the foreground object, corresponding to the inflation points furthest from the actual point cloud's depth-discontinuous edge. When the centerline of the LiDAR beam moves further up, and the beam is entirely on the background object, the returned measurement value is that of the background object.

The mean \(\mu\) varies with the position of the beam's centerline. When the centerline of the beam is tangent to the edge of the foreground object, \(\mu=0\); when the beam as a whole is tangent to the foreground object, \(\mu = \mathbf{e}_i(d_i, \theta)\). Here, \(\mathbf{e}_i\) represents the radius of the circular beam, which is determined by the depth measurement \(d_i\) and the divergence angle \(\theta\) of the LiDAR beam. It is noteworthy that when extracting edges from the intensity image of the LiDAR, the depth-discontinuous edges extracted are the farthest inflation points from the real edge, which are the points where the beam as a whole is tangent to the foreground object. These points conform to a biased Gaussian distribution \((\mu, ^L\mathbf{\Sigma}_{i}) \). We define \(\mathbf{E}_i\) as the bias on the depth-discontinuous edges that we extract. For such inflation points on the depth-discontinuous edges, if treated as actual points on the depth-discontinuous edge, the relation should be 
\begin{align}
^L\mathbf{P}_i - \mathbf{E}_i - {}^L{\mathbf{P}_{i}^{gt}} \sim \mathcal{N}(0, ^L\mathbf{\Sigma}_{i}), 
\end{align}
where
\begin{align*}
\mathbf{E}_i = \mathbf{e}_i{\mathbf{V}}, ~\mathbf{V}=\mathbf{N} \times \mathbf{L},
\end{align*}
the cross product (\(\times\)) operation between \(\mathbf{N}\) and \(\mathbf{L}\) yields vector \(\mathbf{V}\), which is orthogonal to both \(\mathbf{N}\) and \(\mathbf{L}\). \(\mathbf{N}\) is the normal vector of the plane where the edge lies, and \(\mathbf{L}\) is the vector perpendicular from the center of the beam to the actual edge on the surface.

\begin{figure}[htbp] 
\centering 
\includegraphics[width=0.4\columnwidth]{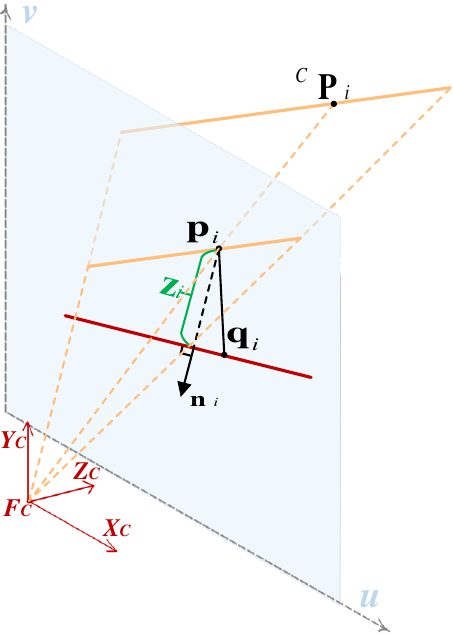} 
\caption{Projection of an edge point from the LiDAR, denoted as  $ ^C\mathbf{P}_i $ in the camera frame, onto the image plane $\mathbf{p}_i $. This includes calculating the residual $\mathbf{z}_i $, associated with the projection. LiDAR edges are marked in blue, and camera edges in red.} 
\label{5} 
\vspace{-0.2cm}
\end{figure}

\subsubsection{Iterative Optimization}\label{optimization} 

Consider $ ^L\mathbf{P}_i $ as an edge point derived from LiDAR point cloud, with the corresponding edge in the imagery characterized by the normal vector $ \mathbf{n}_i $ located within $ \mathbb{S}^1 $, and a relevant point $ \mathbf{q}_i $ in $ \mathbb{R}^2 $ situated on this edge (Section \ref{measurement noises}). Upon the consideration of the noise inherent in $ ^L\mathbf{P}_i$, and upon the projection of this point onto the image plane utilizing validated extrinsic parameters, it is anticipated that it will align precisely with the edge defined by $ (\mathbf{n}_i, \mathbf{q}_i) $ in the image (Eqs. \eqref{eq:projection}, \eqref{eq:2} and Fig. \ref{5}). The optimal observation model is articulated as follows:
\begin{align}
0 = \mathbf{n}_i^T  (\bm{\pi} ( {^{C}_L}\mathbf{T} \left( ^L\mathbf{P}_i - {}^{L}\mathbf{w}_i ) )  - \left( \mathbf{q}_i - {}^{C}\mathbf{w}_i \right) \right)
\label{eq:6},
\end{align}
where the noise components associated with the LiDAR and camera measurements are denoted by $^L\mathbf{w}_i$ and $^C\mathbf{w}_i$, respectively.

The restriction on the extrinsic parameters, imposed by a LiDAR edge point due to its derivation from two distinct points, is demonstrated by Eq. \eqref{eq:6}. Furthermore, Eq. (\ref{eq:6}) formulates a nonlinear relationship pertaining to the measurements of $^L\mathbf{P}_i $, $\mathbf{n}_i$, and $ \mathbf{q}_i$ in relation to the extrinsic calibration ${^C_L}\mathbf{T}$. To address this nonlinear relationship, an iterative methodology is utilized, commencing with the current approximation ${^C_L}\hat{\mathbf{T}}$, and parameterizing ${^C_L}\mathbf{T}$ within the tangent space of ${^C_L}\hat{\mathbf{T}}$ through the \(\boxplus \) operation as delineated in $SE(3)$ \cite{he2021embedding}.
\begin{align}
{^C_L}\mathbf{T} = {^C_L}\hat{\mathbf{T}} \mathop{\boxplus_{SE(3)}} \delta \mathbf{T} \triangleq \text{Exp}(\delta \mathbf{T}) \cdot {^C_L}\hat{\mathbf{T}},\label{eq:guance}
\end{align}


By integrating Eq. (\ref{eq:guance}) into Eq. (\ref{eq:6}) and subsequently approximating the resulting mathematical expression through a first-order term, we get
\begin{align}
    0 \approx \mathbf{z}_i + \mathbf{H}_{\mathbf{T}_i}\delta \mathbf{T}_i + \mathbf{H}_{\mathbf{w}_i}\mathbf{w}_i,
\end{align}
where
\begin{align}
\mathbf{z}_i = & \mathbf{n}_i^T \left(\bm{\pi}\left({^C_L}\mathbf{T} (^L\mathbf{P}_i-\mathbf{E}_i) \right) - \mathbf{q}_i \right) \in \mathbb{R},
\end{align}
the inflation factor \(\mathbf{E}_i\) is delineated in Section \ref{measurement noises}. In the context of depth-continuous edges and intensity-discontinuous edges situated on a plane, \(\mathbf{E}_i\) is regarded as null.

Eq. (\ref{7}) elucidates the constraint conditions that are associated with the alignment of edges. The aggregation of all such edge correspondences facilitates the formulation of a comprehensive constraint, which can subsequently be employed to establish a maximal likelihood problem that concurrently minimizes the variance pertaining to the extrinsic parameters. This problem is then addressed utilizing the Gauss-Newton method for optimization, with the Ceres Solver\footnote[7]{\url{http://www.ceres-solver.org/}} functioning as the computational apparatus to iteratively enhance the extrinsic parameters.

\section{EXPERIMENTS AND RESULTS}

In this section, we validate our calibration approach and compare it with other the state-of-the-art methods on two real-world datasets: the SZU campus dataset and the Yuan's public dataset\cite{pixel-level}. 

\subsection{SZU Campus Dataset}  

\begin{figure}[htbp] 
\centering 
\includegraphics[width=0.9\columnwidth]{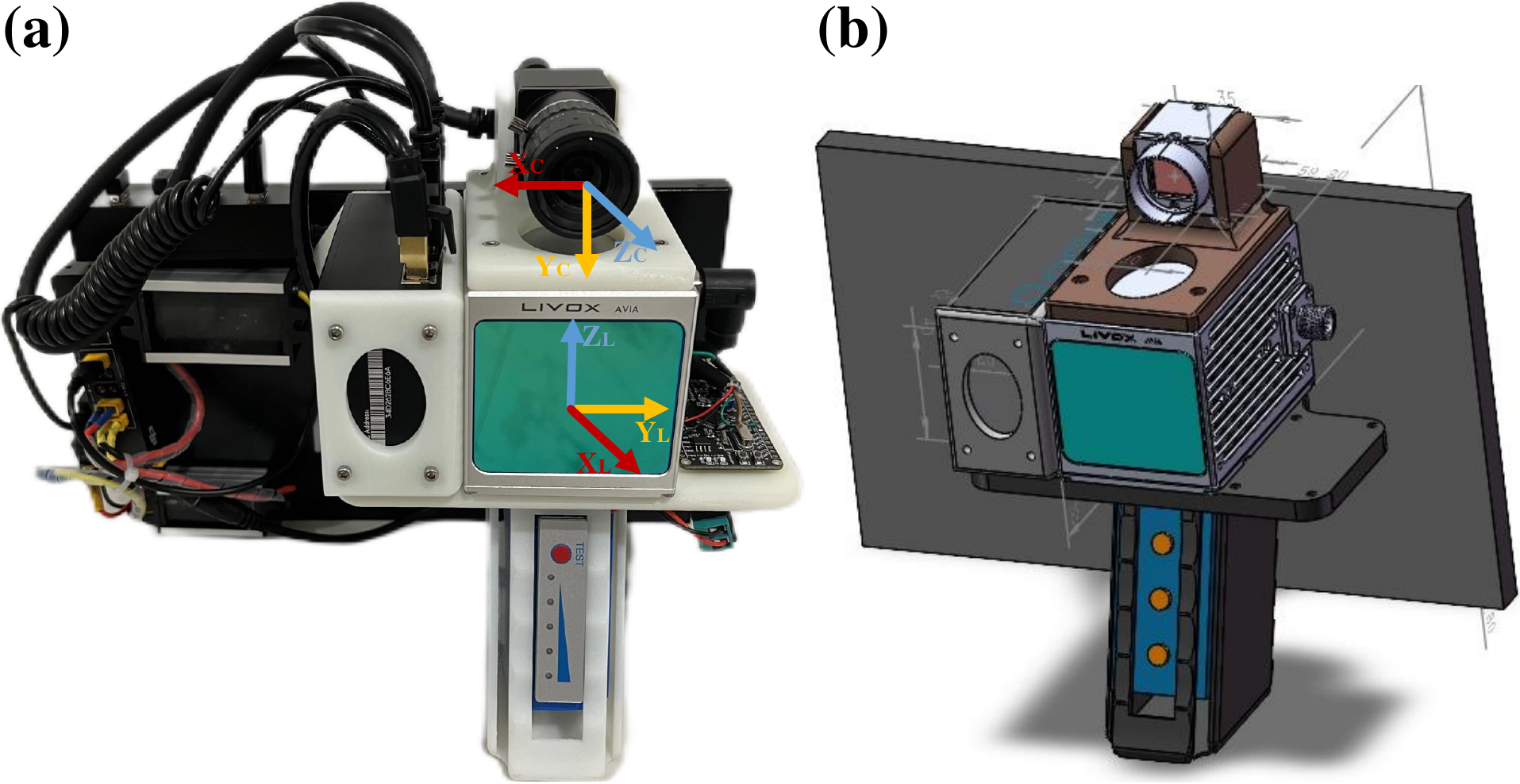} 
\caption{LiDAR-camera sensor suite. (a) Physical assembly of the sensor suite.
(b) CAD model representation of the sensor suite.}
\label{6} 
\vspace{-0.2cm} 
\end{figure}

\begin{figure}[t] 
\centering 
\includegraphics[width=1\columnwidth]{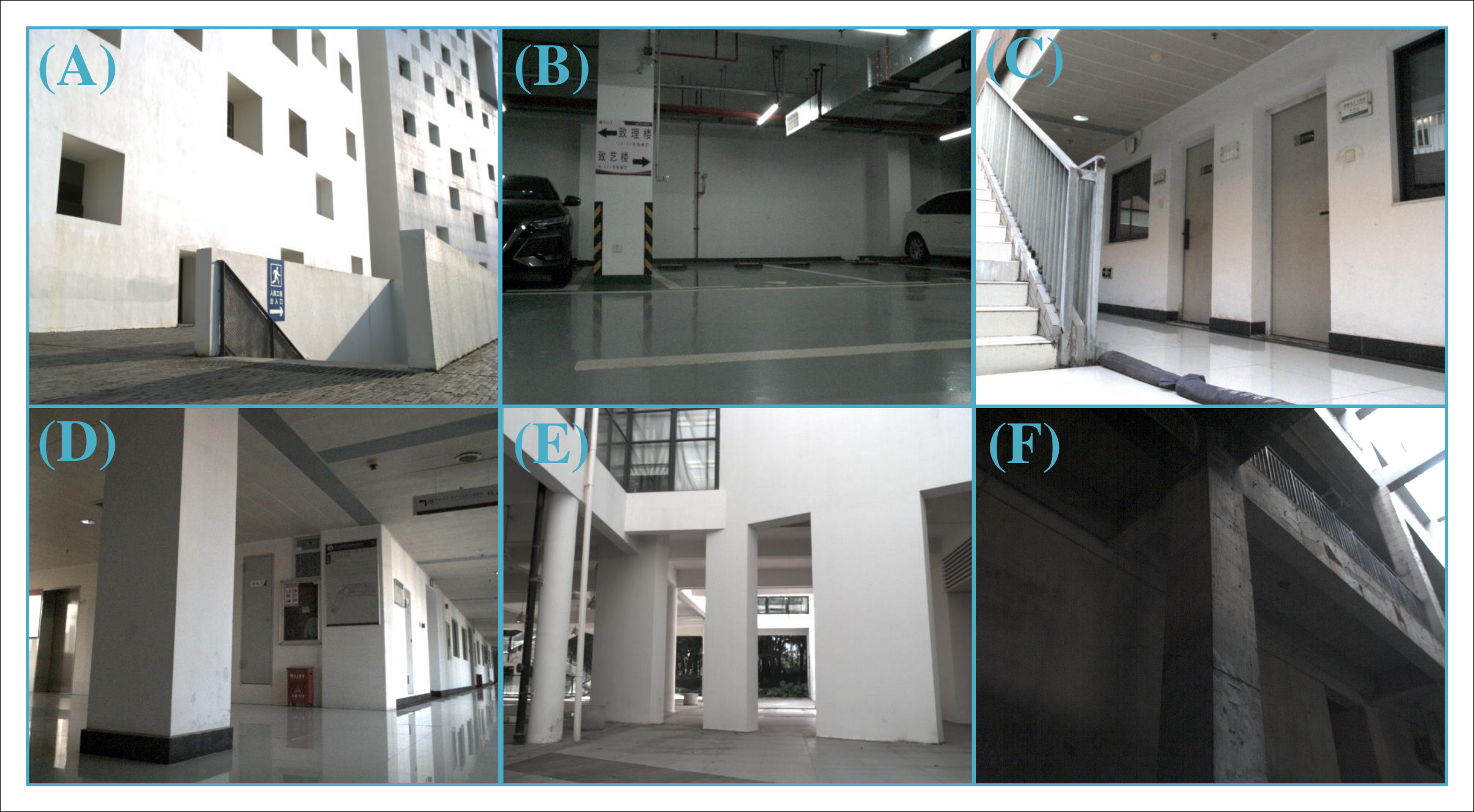}
\caption{Calibration Scenes. (A)--(F) Scene 1--6.} 
\label{7} 
\vspace{-0.5cm} 
\end{figure}

The dataset is collected within the Shenzhen University, employing the sensor suite\footnote[8]{\url{https://github.com/hku-mars/FAST-LIVO}} illustrated in Fig. \ref{6}. We utilized a Livox Avia LiDAR coupled with an industrial-grade camera, the HIKVISION MV-CA013-A0UC\footnote[9]{\url{https://www.hikvisionweb.com/product-category/camera/usb3-0/}}, which boasts a 1.3-megapixel resolution. The camera’s intrinsic parameters are pre-calibrated utilizing the MATLAB Camera Calibration toolbox\footnote[10]{\url{https://ww2.mathworks.cn/help/vision/camera-calibration.html/}}. The LiDAR and camera are statically positioned at the invariant location during data collection. To fully exploit the non-repetitive scanning feature of LiDAR, we accumulate point cloud data over a period of 30 seconds to obtain a dense point cloud. We collect data from six different scenes within the campus (see Fig. \ref{7}), encompassing both indoor (BCD) and outdoor (AEF) settings.

\subsubsection{Robustness and Convergence Validation}
\begin{figure}[htbp] 
\centering 
\includegraphics[width=1\columnwidth]{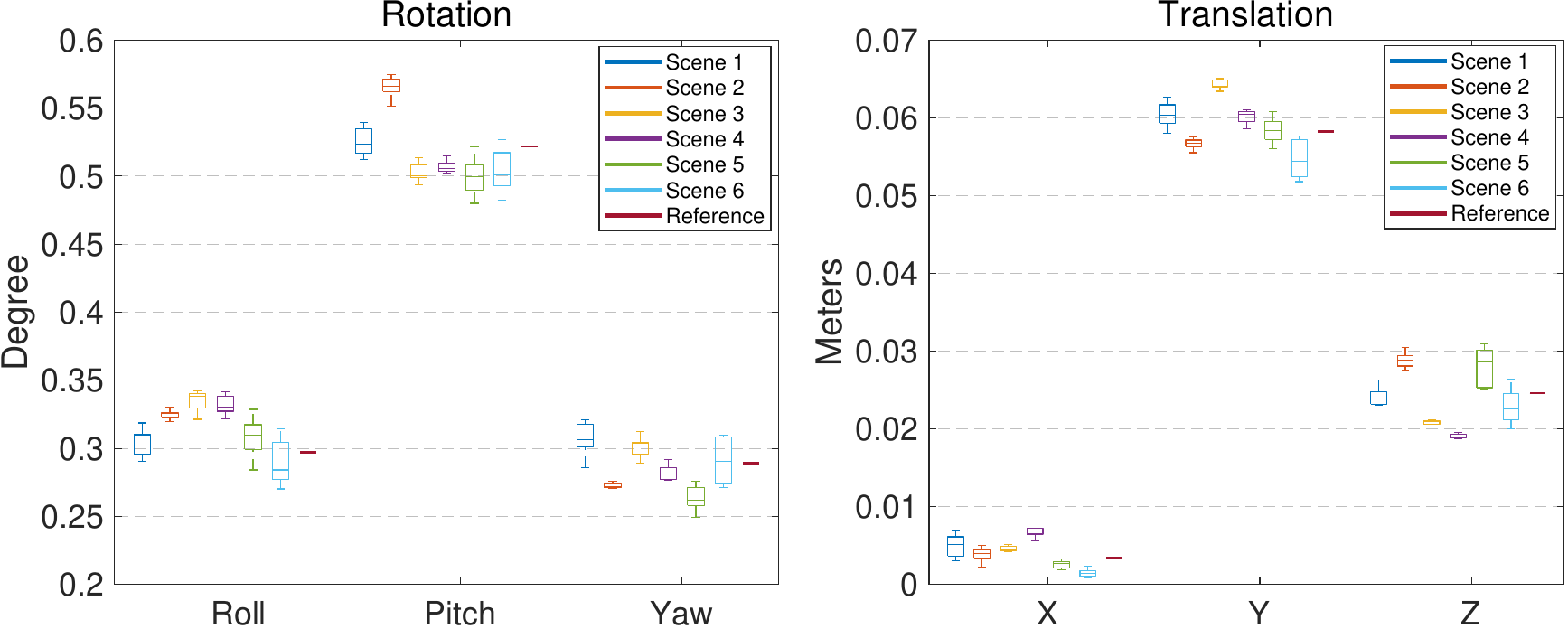} 
\caption{Distribution of converged extrinsic values for all scene settings. The displayed extrinsic has its nominal part removed.} 
\label{8} 
\vspace{-0.3cm}
\end{figure}

To validate the robustness of our algorithm against environmental variations and its accuracy across different scenarios, we conduct tests within the aforementioned six scenes. For each scene, ten sets of trials are performed with varying initial parameters, randomly assigned around the baseline CAD values (rotation \(5^\circ\) and translation 10 cm). Fig. \ref{8} displays the distribution of the extrinsic parameters' results along various axes. It is evident that the extrinsic parameters from different initial values in various scenes converge and approximate the reference values, demonstrating the consistency, robustness, and accuracy of our algorithm in diverse environments.
 
\subsubsection{Comparison Experiments}

For quantitative performance comparison, we calculate the calibration error between the calibrated extrinsic parameters and the ``pseudo" ground-truth values, where the ``pseudo" ground-truth values used for reference are obtained using the \cite{koide2023general} across 20 scenes. The rotation error is given by 
\(
\epsilon_\mathbf{R} = \arccos(\frac{1}{2}(\text{tr}({^{C}_L}\mathbf{R} ({^{C}_L}\mathbf{R}_{gt})^T) - 1)),    
\)
expressed in degrees. The translation error is provided by \( \epsilon_\mathbf{t} = \| {^{C}_L}\mathbf{t} - {^{C}_L}\mathbf{t}_{gt} \| \), expressed in centimeters. The terms \( {^{C}_L}\mathbf{R}_{gt} \) and \( {^{C}_L}\mathbf{t}_{gt} \) denote the reference ``pseudo" ground-truth for rotation and translation, respectively. The calibrated results are denoted by \( {^{C}_L}\mathbf{R} \) and \( {^{C}_L}\mathbf{t} \). We compare our method in the dataset with two open-source the state-of-the-art targetless calibration approaches. The technique in \cite{pixel-level} effectively extracts depth-continuous edges from LiDAR data through voxel segmentation and plane fitting, aligning them with camera edge features to achieve pixel-level calibration. It relies on the presence of depth-continuous edges within the scene, requiring a highly structured calibration environment. Koide \textit{et al.} \cite{koide2023general} developed an open-source LiDAR-camera calibration toolbox that is compatible with various LiDAR and camera models. Their approach also incorporates a direct LiDAR-camera fine registration algorithm based on Normalized Information Distance (NID) and a viewpoint-based hidden point removal algorithm to enhance calibration accuracy. Our method, along with the other two, utilizes identical initial extrinsic parameters. Comparisons are made within corresponding individual scenes. The parameters for each method are tuned to achieve the best results as per the authors' efforts, and the same parameters are used across all scenes for each method.

\begin{table}[htbp]
  \centering
  \caption{Rotation Errors (Degrees) and Translation Errors (Centimeters)}
  \label{tab:1}
  \scriptsize 
  \setlength\tabcolsep{1.8pt} 
  \begin{tabular}{@{}cccccccc@{}}
    \toprule
    & Scene 1 & Scene 2 & Scene 3 & Scene 4 & Scene 5 & Scene 6 &Average \\ 
    \midrule
    \textbf{Ours} & \textbf{0.18/1.53} & \textbf{0.08/1.73} & 0.20/\textbf{1.62} & \textbf{0.16/1.66} & \textbf{0.23/1.61} & \textbf{0.24/1.43} & \textbf{0.18/1.60}\\
    Yuan's & 0.22/2.55 & 2.31/43.21 & 0.48/5.70 & 0.38/3.80 & 0.27/7.92 & 0.54/5.88 & 0.70/11.51\\
    Koide's & 0.44/21.73 & 1.17/75.55 & \textbf{0.18}/2.39 & 0.35/7.07 & 0.13/3.57 & 0.56/11.75 & 0.47/20.34\\
    \bottomrule
  \end{tabular}
\vspace{-0.2cm}
\end{table}

\begin{figure*}[htbp] 
\centering
\includegraphics[width=2\columnwidth]{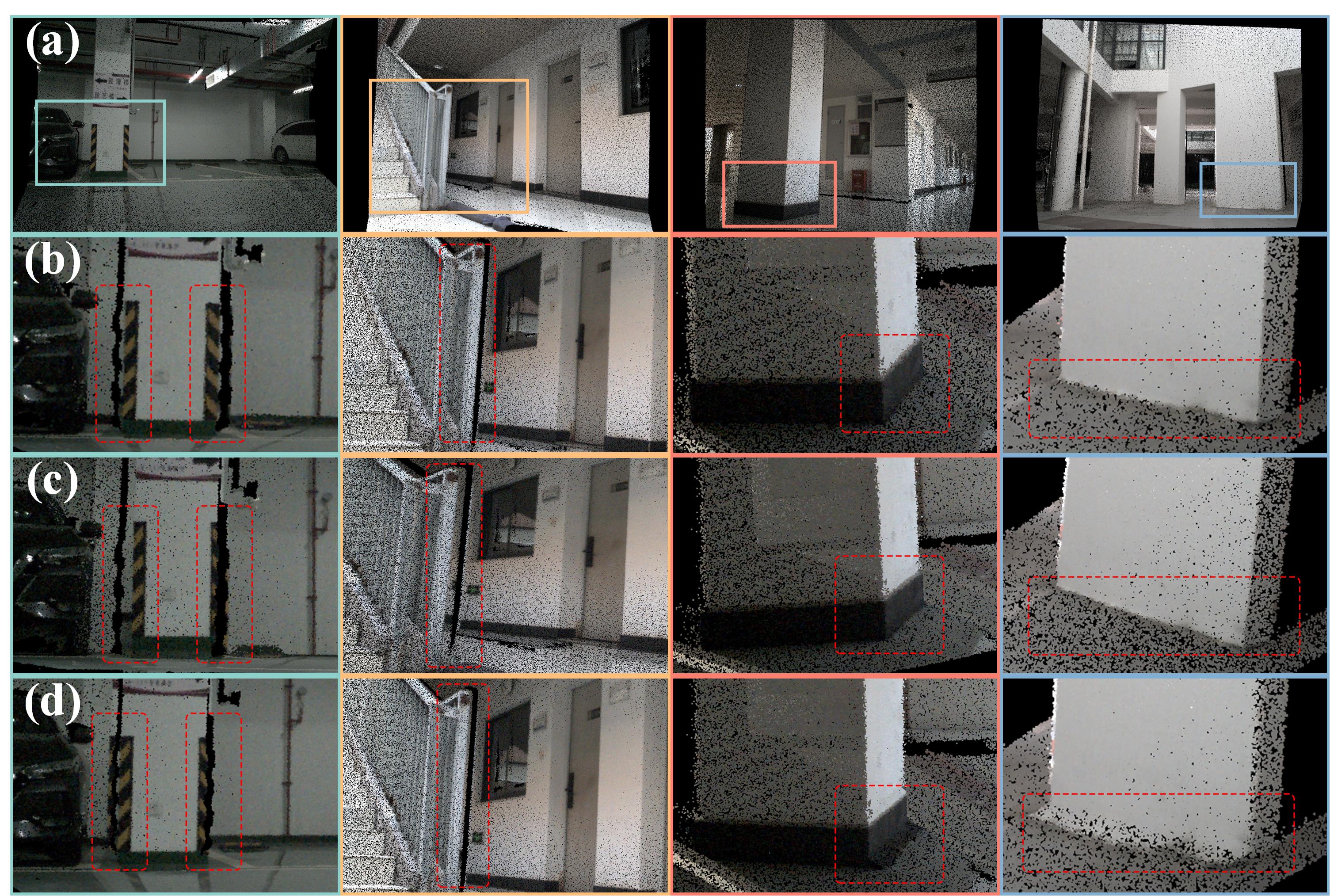} 
\vspace{-0.2cm}
\caption{(a) Colorized point clouds using ``pseudo" ground-truth calibration parameters, with the area within the square frame designated for comparison. (b) The result of colorizing with our calibration result. (c) The result of colorizing with Yuan's calibration result. (d) The result of colorizing with Koide's calibration result. The red boxes emphasize areas with noticeable differences in color alignment.} 
\label{colorpcd}
\vspace{-0.3cm}
\end{figure*}

\begin{figure*}[htbp] 
\centering
\includegraphics[width=2\columnwidth]{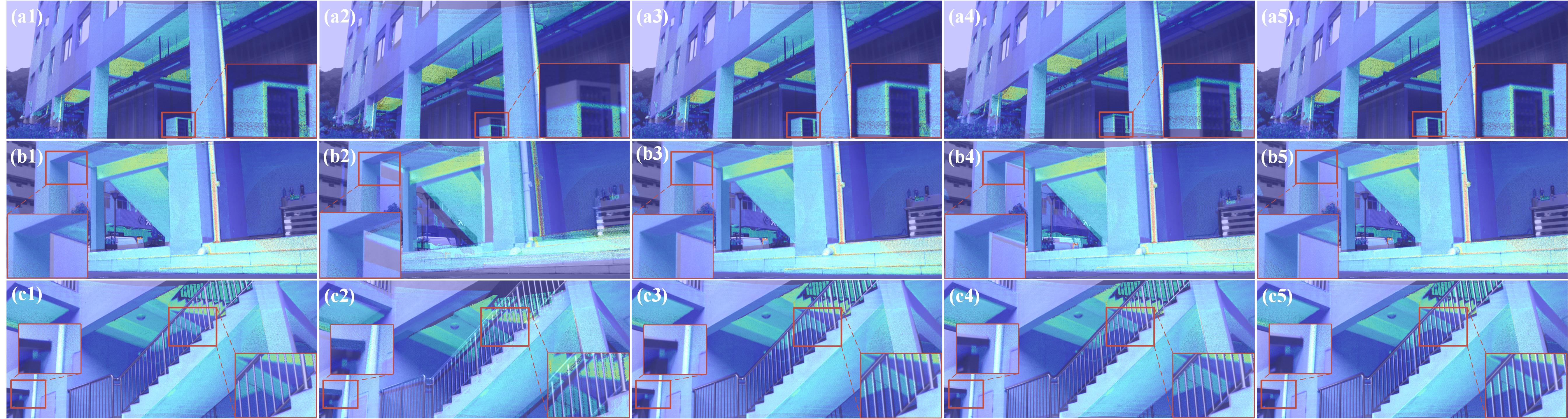} 
\vspace{-0.2cm}
\caption{Using three Scenes from Yuan's public dataset \cite{pixel-level}, we evaluated the performance of our method against Yuan's and Koide's methods. Images show LiDAR data fused with camera imagery, calibrated extrinsically, with point intensities represented by a Jet colormap. (a1), (b1), and (c1) present our method results calibrated solely within the respective single scene. (a2), (b2), and (c2) present Yuan's method results calibrated solely within the respective single scene. (a3), (b3), and (c3) display Yuan's method results calibrated using all three scenes together. (a4), (b4), and (c4) present Koide's method results calibrated solely within the respective single scene. (a5), (b5), and (c5) display Koide's method results calibrated using all three scenes together.} 
\label{9}
\vspace{-0.3cm}
\end{figure*}

The initial rotation and translation errors are 0.91°/8.49cm, respectively. The experimental outcomes are encapsulated in Table \ref{tab:1}. Our method demonstrates precise calibration results across six scenes. In contrast, Yuan’s method failed in Scene 2, while Koide’s method failed in Scenes 1 and 2, underscoring the superior robustness of our approach. Even in scenes where other methods are successful, ours achieves higher accuracy, although there is a slight increase in rotational error in Scene 3 compared to Koide’s method. The exemplary performance of our method is attributed to the use of a diversified feature set for matching, along with consideration of the LiDAR beam uncertainties. In contrast, Koide’s method relies on feature extraction from visual and intensity images, which poses significant challenges under suboptimal conditions, and Yuan’s method requires a rich presence of depth-continuous edges in the scene.

Beyond quantitative comparison, we conducted a further assessment of calibration quality. Based on the calibration results from Table \ref{tab:1}, we colorized the LiDAR point clouds and selected Scenes 2–4, which have a higher color contrast between foreground and background objects, to enhance visual comparison (see Fig. \ref{colorpcd}). The results demonstrate that our method consistently produces accurate and visually coherent colorizations across all scenes. In contrast, methods proposed by Yuan and Koide show noticeable color misalignments in certain scenarios. This visual evaluation further confirms the superiority of our method in terms of calibration accuracy and reliability compared to other state-of-the-art approaches.

\subsection{Yuan's Public Dataset}

We conduct comparative tests on Yuan's public dataset\cite{pixel-level} to further validate our algorithm. This dataset is collected using a sensor suite composed of a Livox Avia LiDAR and an Intel Realsense-D435i camera\footnote[11]{\url{ https://www.intelrealsense.com/depth-camera-d435i}}. We select three scenes from the dataset for experimental verification. Our method utilizes calibration within individual scenes exclusively, while the approaches by Yuan and Koide considered both individual scenes and the integration of multiple scenes. As shown in Fig. \ref{9}, the results demonstrate that our method significantly outperforms the other two methods under the same single-scene conditions and achieves or exceeds the accuracy of their multi-scene calibration results. This highlights the high precision of our method in single-scene scenarios and confirms its robustness in challenging environments.

\section{CONCLUSIONS}

In this study, we introduce an innovative targetless automatic extrinsic calibration method for LiDAR and camera systems that requires only a single data collection pass. Our approach, integrating multi-feature edge extraction techniques with a LiDAR beam model, effectively addresses the issue of edge inflation caused by LiDAR. This method surpasses existing the state-of-the-art targetless techniques in precision and robustness, proving its efficacy even in challenging scenarios where traditional methods fall short.

	\section*{Acknowledgment}
	The authors gratefully acknowledge James Lam for his guidance and support in improving the writing of this work.

\addtolength{\textheight}{-12cm}   









\bibliographystyle{IEEEtran}
\bibliography{paper}

\begin{thebibliography}{10}
\providecommand{\url}[1]{#1}
\csname url@samestyle\endcsname
\providecommand{\newblock}{\relax}
\providecommand{\bibinfo}[2]{#2}
\providecommand{\BIBentrySTDinterwordspacing}{\spaceskip=0pt\relax}
\providecommand{\BIBentryALTinterwordstretchfactor}{4}
\providecommand{\BIBentryALTinterwordspacing}{\spaceskip=\fontdimen2\font plus
\BIBentryALTinterwordstretchfactor\fontdimen3\font minus
  \fontdimen4\font\relax}
\providecommand{\BIBforeignlanguage}[2]{{%
\expandafter\ifx\csname l@#1\endcsname\relax
\typeout{** WARNING: IEEEtran.bst: No hyphenation pattern has been}%
\typeout{** loaded for the language `#1'. Using the pattern for}%
\typeout{** the default language instead.}%
\else
\language=\csname l@#1\endcsname
\fi
#2}}
\providecommand{\BIBdecl}{\relax}
\BIBdecl

\bibitem{ORB-SLAM2}
R.~Mur-Artal and J.~D. Tard{\'o}s, ``{ORB-SLAM2: An open-source SLAM system for
  monocular, stereo, and RGB-D cameras},'' \emph{IEEE Transactions on
  Robotics}, vol.~33, no.~5, pp. 1255--1262, 2017.

\bibitem{xu2021fast}
W.~Xu and F.~Zhang, ``{FAST-LIO: A Fast, Robust {LiDAR}-Inertial Odometry
  Package by Tightly-Coupled Iterated {Kalman} Filter},'' \emph{IEEE Robotics
  and Automation Letters}, vol.~6, no.~2, pp. 3317--3324, 2021.

\bibitem{point-lio}
D.~He, W.~Xu, N.~Chen, F.~Kong, C.~Yuan, and F.~Zhang, ``{Point-LIO}: {Robust}
  high-bandwidth light detection and ranging inertial odometry,''
  \emph{Advanced Intelligent Systems}, p. 2200459, 2023.

\bibitem{voxel-map}
C.~Yuan, W.~Xu, X.~Liu, X.~Hong, and F.~Zhang, ``Efficient and probabilistic
  adaptive voxel mapping for accurate online {LiDAR} odometry,'' \emph{IEEE
  Robotics and Automation Letters}, vol.~7, no.~3, pp. 8518--8525, 2022.

\bibitem{peng2023bevsegformer}
L.~Peng, Z.~Chen, Z.~Fu, P.~Liang, and E.~Cheng, ``{BEVSegFormer: Bird's eye
  view semantic segmentation from arbitrary camera rigs},'' in
  \emph{Proceedings of the IEEE/CVF Winter Conference on Applications of
  Computer Vision}, 2023, pp. 5935--5943.

\bibitem{ding2023lenet}
B.~Ding, ``{LENet: Lightweight and efficient LiDAR semantic segmentation using
  multi-scale convolution attention},'' \emph{arXiv preprint arXiv:2301.04275},
  2023.

\bibitem{zhu2021camvox}
Y.~Zhu, C.~Zheng, C.~Yuan, X.~Huang, and X.~Hong, ``{Camvox: A low-cost and
  accurate {LiDAR}-assisted visual SLAM system},'' in \emph{Proceedings of 2021
  IEEE International Conference on Robotics and Automation (ICRA)}, 2021, pp.
  5049--5055.

\bibitem{lin2021r}
J.~Lin, C.~Zheng, W.~Xu, and F.~Zhang, ``{R2LIVE: A Robust, Real-Time,
  {LiDAR}-Inertial-Visual Tightly-Coupled State Estimator and Mapping},''
  \emph{IEEE Robotics and Automation Letters}, vol.~6, no.~4, pp. 7469--7476,
  2021.

\bibitem{zheng2022fast}
C.~Zheng, Q.~Zhu, W.~Xu, X.~Liu, Q.~Guo, and F.~Zhang, ``{FAST-LIVO: Fast and
  Tightly-coupled Sparse-Direct LiDAR-Inertial-Visual Odometry},'' in
  \emph{Proceedings of 2022 IEEE/RSJ International Conference on Intelligent
  Robots and Systems (IROS)}, 2022, pp. 4003--4009.

\bibitem{zheng2024fast}
C.~Zheng, W.~Xu, Z.~Zou, T.~Hua, C.~Yuan, D.~He, B.~Zhou, Z.~Liu, J.~Lin,
  F.~Zhu \emph{et~al.}, ``Fast-livo2: Fast, direct lidar-inertial-visual
  odometry,'' \emph{arXiv preprint arXiv:2408.14035}, 2024.

\bibitem{lin2022r}
J.~Lin and F.~Zhang, ``{R3LIVE: A robust, real-time, RGB-colored,
  LiDAR-inertial-visual tightly-coupled state estimation and mapping
  package},'' in \emph{Proceedings of 2022 International Conference on Robotics
  and Automation (ICRA)}, 2022, pp. 10\,672--10\,678.

\bibitem{beltran2022automatic}
J.~Beltr{\'a}n, C.~Guindel, A.~de~la Escalera, and F.~Garc{\'\i}a, ``{Automatic
  extrinsic calibration method for LiDAR and camera sensor setups},''
  \emph{IEEE Transactions on Intelligent Transportation Systems}, vol.~23,
  no.~10, pp. 17\,677--17\,689, 2022.

\bibitem{cui2020acsc}
J.~Cui, J.~Niu, Z.~Ouyang, Y.~He, and D.~Liu, ``{ACSC: Automatic calibration
  for non-repetitive scanning solid-state LiDAR and camera systems},''
  \emph{arXiv preprint arXiv:2011.08516}, 2020.

\bibitem{Zhou_Li_Kaess_2018}
L.~Zhou, Z.~Li, and M.~Kaess, ``{Automatic extrinsic calibration of a camera
  and a 3D LiDAR using line and plane correspondences},'' in \emph{Proceedings
  of 2018 IEEE/RSJ International Conference on Intelligent Robots and Systems
  (IROS)}, Oct 2018.

\bibitem{Chen_Liu_Liang_Zhang_Hyyppa_Chen_2020}
S.~Chen, J.~Liu, X.~Liang, S.~Zhang, J.~Hyyppa, and R.~Chen, ``{A novel
  calibration method between a camera and a 3D LiDAR with infrared images},''
  in \emph{Proceedings of 2020 IEEE International Conference on Robotics and
  Automation (ICRA)}, May 2020.

\bibitem{pixel-level}
C.~Yuan, X.~Liu, X.~Hong, and F.~Zhang, ``Pixel-level extrinsic self
  calibration of high resolution {LiDAR} and camera in targetless
  environments,'' \emph{IEEE Robotics and Automation Letters}, vol.~6, no.~4,
  pp. 7517--7524, 2021.

\bibitem{Chen_Li_Zhang_Wu_Wang_2023}
F.~Chen, L.~Li, S.~Zhang, J.~Wu, and L.~Wang, ``{PBACalib: Targetless extrinsic
  calibration for high-resolution LiDAR-camera system based on
  plane-constrained bundle adjustment},'' \emph{IEEE Robotics and Automation
  Letters}, pp. 304--–311, 2023.

\bibitem{Xie_Shao_Guli_Li_Wang_2018}
Y.~Xie, R.~Shao, P.~Guli, B.~Li, and L.~Wang, ``{Infrastructure based
  calibration of a multi-camera and multi-LiDAR system using apriltags},'' in
  \emph{Proceedings of 2018 IEEE Intelligent Vehicles Symposium (IV)}, Jun
  2018.

\bibitem{Kummerle_Kuhner_2020}
J.~Kummerle and T.~Kuhner, ``{Unified intrinsic and extrinsic camera and LiDAR
  calibration under uncertainties},'' in \emph{Proceedings of 2020 IEEE
  International Conference on Robotics and Automation (ICRA)}, May 2020.

\bibitem{Park_Yun_Won_Cho_Um_Sim_2014}
Y.~Park, S.~Yun, C.~Won, K.~Cho, K.~Um, and S.~Sim, ``{Calibration between
  color camera and 3D LiDAR instruments with a polygonal planar board},''
  \emph{Sensors}, pp. 5333–--5353, Mar 2014.

\bibitem{chen2022pbacalib}
F.~Chen, L.~Li, S.~Zhang, J.~Wu, and L.~Wang, ``{PBACalib: Targetless extrinsic
  calibration for high-resolution LiDAR-camera system based on
  plane-constrained bundle adjustment},'' \emph{IEEE Robotics and Automation
  Letters}, vol.~8, no.~1, pp. 304--311, 2022.

\bibitem{li2023joint}
L.~Li, H.~Li, X.~Liu, D.~He, Z.~Miao, F.~Kong, R.~Li, Z.~Liu, and F.~Zhang,
  ``{Joint intrinsic and extrinsic LiDAR-camera calibration in targetless
  environments using plane-constrained bundle adjustment},'' \emph{arXiv
  preprint arXiv:2308.12629}, 2023.

\bibitem{Liu_Yuan_Zhang_2022}
X.~Liu, C.~Yuan, and F.~Zhang, ``{Targetless extrinsic calibration of multiple
  small FoV LiDARs and cameras using adaptive voxelization},'' \emph{IEEE
  Transactions on Instrumentation and Measurement}, pp. 1–--12, Jan 2022.

\bibitem{miao2023coarse}
Z.~Miao, B.~He, W.~Xie, W.~Zhao, X.~Huang, J.~Bai, and X.~Hong,
  ``{Coarse-to-fine hybrid 3D mapping system with co-calibrated omnidirectional
  camera and non-repetitive LiDAR},'' \emph{IEEE Robotics and Automation
  Letters}, vol.~8, no.~3, pp. 1778--1785, 2023.

\bibitem{koide2023general}
K.~Koide, S.~Oishi, M.~Yokozuka, and A.~Banno, ``{General, single-shot,
  target-less, and automatic LiDAR-camera extrinsic calibration toolbox},''
  \emph{arXiv preprint arXiv:2302.05094}, 2023.

\bibitem{Pandey_McBride_Savarese_Eustice_2022}
G.~Pandey, J.~McBride, S.~Savarese, and R.~Eustice, ``{Automatic targetless
  extrinsic calibration of a 3D Lidar and camera by maximizing mutual
  information},'' in \emph{Proceedings of the AAAI Conference on Artificial
  Intelligence}, Jun 2022, pp. 2053–--2059.

\bibitem{Sarlin_DeTone_Malisiewicz_Rabinovich_2020}
P.-E. Sarlin, D.~DeTone, T.~Malisiewicz, and A.~Rabinovich, ``{SuperGlue:
  Learning feature matching with graph neural networks},'' in \emph{Proceedings
  of 2020 IEEE/CVF Conference on Computer Vision and Pattern Recognition
  (CVPR)}, Jun 2020.

\bibitem{Taylor_Nieto_2016}
Z.~Taylor and J.~Nieto, ``Motion-based calibration of multimodal sensor
  extrinsics and timing offset estimation,'' \emph{IEEE Transactions on
  Robotics}, pp. 1215--–1229, Oct 2016.

\bibitem{Liu_Zhang_2021}
X.~Liu and F.~Z. Zhang, ``{Extrinsic calibration of multiple LiDARs of small
  FoV in targetless environments},'' \emph{IEEE Robotics and Automation
  Letters}, pp. 2036--–2043, Apr 2021.

\bibitem{canny1986computational}
J.~Canny, ``A computational approach to edge detection,'' \emph{IEEE
  Transactions on Pattern Analysis and Machine Intelligence}, no.~6, pp.
  679--698, 1986.

\bibitem{fast-lio2}
W.~Xu, Y.~Cai, D.~He, J.~Lin, and F.~Zhang, ``{FAST-LIO2: Fast Direct
  Lidar-Inertial Odometry},'' \emph{IEEE Transactions on Robotics}, vol.~38,
  no.~4, pp. 2053--2073, 2022.

\bibitem{yuan2021arxiv}
C.~Yuan, X.~Liu, X.~Hong, and F.~Zhang, ``Pixel-level extrinsic self
  calibration of high resolution lidar and camera in targetless environments,''
  \emph{arXiv preprint arXiv:2103.01627}, 2021.

\bibitem{he2021embedding}
D.~He, W.~Xu, and F.~Zhang, ``Embedding manifold structures into kalman
  filters,'' \emph{arXiv preprint arXiv:2102.03804}, 2021.

\end{thebibliography}

\end{document}